# DeepDeath: Learning to Predict the Underlying Cause of Death with Big Data*

Hamid Reza Hassanzadeh, *Student Member, IEEE*, Ying Sha, and May D. Wang, *Senior Member, IEEE*

*Abstract—* **Multiple cause-of-death data provides a valuable source of information that can be used to enhance health standards by predicting health related trajectories in societies with large populations. These data are often available in large quantities across U.S. states and require Big Data techniques to uncover complex hidden patterns. We design two different classes of models suitable for large-scale analysis of mortality data, a Hadoop-based ensemble of random forests trained over N-grams, and the DeepDeath, a deep classifier based on the recurrent neural network (RNN). We apply both classes to the mortality data provided by the National Center for Health Statistics and show that while both perform significantly better than the random classifier, the deep model that utilizes long short-term memory networks (LSTMs), surpasses the N-gram based models and is capable of learning the temporal aspect of the data without a need for building ad-hoc, expert-driven features.**

## I. Introduction

Many of the scientific discussions and studies in biomedical and healthcare domains address tasks whose end goal is to prevent death or diseases. Since the emergence of the big data science, numerous machine learning based techniques and technologies have been proposed and applied to improve human health by solving different computational challenges that we face today. A less obvious question, that remains to be extensively explored by researchers, is whether Big Data science can contribute to our understanding of factors leading to death or diseases, via analysis of multiple-cause mortality data. In fact it is widely believed that counting the dead is a significant investment to reduce the premature mortality [2]. There have been a number of studies that have proven to offer profound impacts on our understanding of the major causes of death using the statistical analysis of recorded death data. In light of these studies, we were interested to investigate the feasibility of this emerging field in learning hidden complex patterns that are available in the haystack of mortality datasets.

Multiple causes of death data provide a valuable source of information that have been used to analyze the death event in chronic diseases such as the HIV [3, 4] and the lung disease [5], to identify problems with the process of coding/recording cause-of-death information [6]. Moreover, these data can be potentially used in analysis of disease diffusion for controlling plagues and other epidemics and may provide a better understanding of multi-morbid associations between conditions leading to death. As such, designing advanced analytics pipelines for discovering descriptive statistics and trajectories is highly crucial. While the sheer amount of available data collected from the registered death certificates makes it amenable to Big Data analysis, it poses some key challenges at the same time. In particular, the multiple-cause-of-death data are unstructured and are often inaccurate and noisy. Moreover, the high number of ICD-9/10 mortality codes makes analysis of multiple-cause associations even more challenging. These altogether, call for advanced techniques for mining large datasets of unstructured, high dimensional, and noisy structure.

Despite the importance of the subject, only a handful of studies have so far conducted research seeking to relate multiple causes of death to each other or to other factors. These studies are often limited to classical statistical methods (measures) that do not scale up efficiently and are put into four major categories [7]: 1) Univariate measures, consisting of counts and frequencies, 2) cross-tabular measures, which incorporate variables that identify the roles (e.g. contributory, non-contributory, complication and underlying) associated with multiple death causes, 3) measures of association, in which some measure of multiple mentions of a cause is related to some measure of mentions of the underlying cause; and finally, 4) derived measures, where univariate measures such as multiple-cause rates are integrated to build higher order models.

In this study, we present an exploratory analysis that is well positioned in a fifth group, by building upon both the third and the fourth abovementioned categories, and utilizing advanced machine learning approaches. More specifically, we propose two different classes of models for large-scale analysis of data, namely, shallow learners to learn uni/bi-gram features derived from the multiple-cause data that run over Hadoop framework using the MapReduce programming model, and a deep recurrent neural network that learns the temporal dynamics of the event chains efficiently. The rest of the paper is organized as follows. In section II we detail the data format as well as the

*This work was supported in part by grants from the US Department of Health and Human Services (HHS) Centers for Disease Control and Prevention (CDC) HHSD2002015F62550B, National Science Foundation Award NSF1651, and Microsoft Research and Hewlett Packard. This article does not reflect the official policy or opinions of the CDC, NSF, or the US Department of HHS and does not constitute an endorsement of the individuals or their programs.

H. R. Hassanzadeh is with the Department of Computational Science and Engineering, Georgia Institute of Technology, Atlanta, GA 30332 USA. (email: hassanzadeh@gatech.edu).

Y. Sha is with the Department of Biology, Georgia Institute of Technology, Atlanta, GA 30332 USA. (email: ysha8@gatech.edu)

M. D. Wang is with the Department of Biomedical Engineering, Georgia Institute of Technology and Emory University and the School of Electrical and Computer Engineering, Georgia Institute of Technology, Atlanta, GA 30332 USA (corresponding author, phone: 404-385-2954; e-mail: maywang@bme.gatech.edu).

challenges that we face when dealing with it. Then we describe the shallow learners that are designed to work over the Hadoop framework. We also present our proposed deep model in the same section and our motivation to resort to deep learning. Next, in section III, we compare the accuracy of each model when applied to a large dataset show that our deep pipeline outperforms the baselines we design by utilizing its ability to learn the temporal aspect of the data and finally in section IV, we conclude the paper and shed light on future directions that we would like to pursue.

## II. MATERIALS AND METHODS

### A. Data source

Civil registration systems collect death information from deceased persons in form of death certificates, based on a standard format that is designed by the World Health Organization (WHO) [8]. The section that is of most interest to public health researchers is the cause-of-death section, which has to be completed by a medical certifier. An ideal person to complete a death certificate is the attending physician, who has sufficient clinical expertise and judgement on the occurred death. However, in case the manner of death is unnatural or unplanned, a medical examiner or a coroner can also fill in the death certificate. The cause-of-death section is divided into two parts (see Figure 1). Part I lists the causal chain of conditions directly leading to the death in reverse chronological order, and part II includes the conditions that contributes but not directly leads to death [9]. The conditions in part II are not ordered in time.

We used the 2015 mortality data published by the United States National Center for Health Statistics (NCHS) which consists of over 2.7 million deaths recorded in U.S. during the year 2015. Because we were interested in the temporal information that is available in the multiple causes of death, we filtered out the conditions listed in part II as well as the underlying causes if listed in this part. We also removed that records with attributed unnatural underlying causes of death such as suicide. Based on the 113 recode of the underlying causes, these are assigned to codes 111-113. Moreover, we excluded the underlying causes that appear less than 1000 times throughout the whole dataset leaving us 67 recoded classes for the underlying causes. Our goal in this study was to predict the class attributed to each case given the multiple cause conditions listed for him/her. We divided the resulting set into a training set and a test set comprising 70% and 20% of the samples, respectively.

### B. Using n-Gram Models to Learn Associations Between Multiple Causes

To construct our baseline models, we derived the n-gram features where an n-gram is a defined as an n-tuple consisting of *n* consecutive tokens within sequential data [10] (in our case, causes listed in part I of the death certificates). N-gram based models have been widely used in natural language processing [11-13] and bioinformatics [14, 15] due to their performance and ease of implementation. In this study, we only use uni-gram features and bi-gram features. Despite the fact that higher order n-grams (e.g. tri-grams) can provide more expressiveness and capture more context from the data, they make the models prone to overfitting due to an exponential increase in the number of possible features, which also makes training the resulting model computationally infeasible, therefore, in this study, we only included uni-gram and bi-gram features.

Figure 1. An screenshot of the multiple cause section in a death certificate [1]

Once the n-gram features derived, we used random forest (RF) to train over the extremely large and sparse matrix of features. RF is an efficient model for dealing with sparse features, however, it does not properly scale up to fit the size of NCHS datasets, therefore, we implemented our model over the Hadoop framework. Hadoop is a distributed storage and computing framework inspired by Google File System (GFS) and Google MapReduce [16]. It uses a reliable storage mechanism, the so called Hadoop Distributed File System (HDFS) for fast storage and retrieval of large size datasets. Hadoop can efficiently utilize commodity compute nodes, that are distributed in a network, to split the tasks into smaller sub-tasks and perform the analysis on the smaller chunks of data through MapReduce, an efficient model that runs on HDFS file system. To train our baseline models over the complete dataset, we used Hadoop streaming and used the python Scikit-learn library to train multiple random forests on subsampled data provided by the mapping stage. Once multiple random forests were trained by the reducer jobs, we aggregate them through majority voting using the scores that each model predict on the test set.

### C. DeepDeath

Deep learning is an emerging technology that is now being deployed in a wide range of domains including Biomedical areas [17-19] due its success in improving the previously recorded state-of-the-art performance measures. Deep learning is now becoming an indispensable part of any winning model in today's complex computational challenges. Long short-term memory networks (LSTMs) [20] are an important class of deep architectures that are able to capture the temporal dynamics of sequential data. LSTM networks have recently proved to outperform surprisingly well in many traditional sequence learning algorithms, such as the hidden Markov models (HMMs) and conditional random fields (CRF), in supervised settings where significant amount of labeled data is available. As such, LSTMs fit the framework of the task we would like to solve, namely, using the timed causes of death events that contributed to the final event, death, to predict the underlying cause that initiated these events in the first place. These models have recently attracted significant attention and interest due to the notable speed-ups grained from utilizing advanced graphics processing units (GPUs) for computational tasks where simple matrix operations are massively repeated to generate the outcome of an algorithm.

Figure 2. Internal Design of an LSTM Module

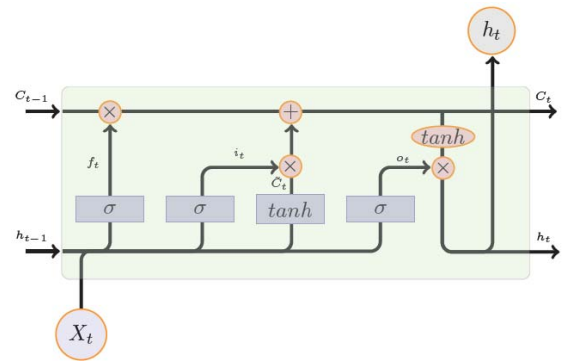

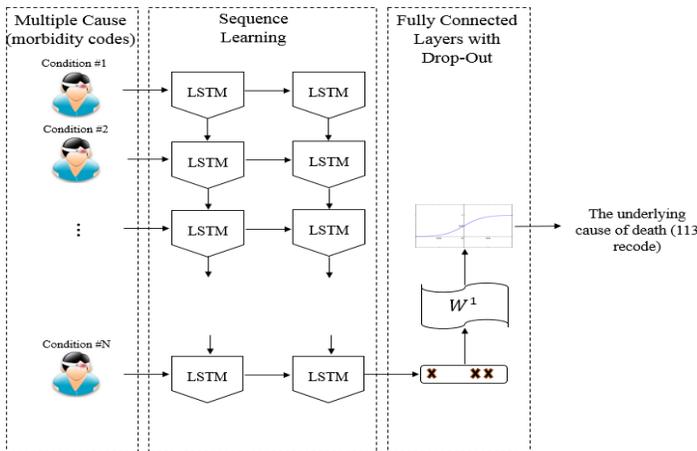

Figure 3. The block diagram of DeepDeath

Figure 2 depicts the internal structure of an LSTM block. It is composed of a number of gates that that control the flow information into and out of each LSTM memory. These gates are programmable in a differentiable way rendering them amenable to any gradient-based optimization technique. In other words, through the training process, we teach these gates what type of information is useful for predicting the target in future and hence should be passed by.

Figure 3 shows the block diagram of DeepDeath. We use a two-layer LSTM network to learn the hidden patterns within the multiple-cause sequences. While each LSTM block in the second layer can potentially generate an output, we only use the output generated by the last block as it can readily integrate all the past history into an abstract representation of the input sequence. We also used drop-out regularization on top of the output of the last block as it has been shown to lead to better a generalization. Once the intermediate features generated, we feed them into a fully connected layer followed by a SoftMax layer to generate the log probability of the sample belonging to each underlying cause class. As a pre-processing step to generate data suitable for DeepDeath, we divided each code into three parts, the group letter, the major code and the etiology and used one-hot coding to represent each part and concatenated the resulting binary codes into a long binary. That way the integrity of the codes is not lost by blindly converting every condition into a single code.

## III. RESULTS

We trained three baseline methods which differ from each other by the pool of the n-gram features. First, a model was trained over 5000 unigram features that appeared the most across all samples. Then, to take advantage of the temporal nature of the inputs, we trained a similar model but over 5000 bi-gram features, and finally, we trained a separate model which include all features from both these models. For the proposed model, we empirically chose sub-optimal hyper-parameters and model architecture that work best on the training set. In particular, we selected the memory size of each LSTM block to be 30 and the drop-out ratio to be 0.1. We trained the network over Tesla C2050 GPUs with RMSProp, an efficient optimization technique for training deep models, for 40 training epochs and a learning rate of 0.003.

We evaluated the DeepDeath as well as the baselines over the randomly compiled test set. Table 1 compares the accuracies achieved for each model. According to the table, all the four models result in an accuracy significantly higher than the accuracy derived from the random classifier (1.49%). While the uni-gram features turn out to be more informative than the bi-gram features, interestingly, the integration of these two results in a 2.56% improvement in the classification performance, suggesting that the temporal nature of the data convey some useful information that cannot be otherwise captured. Finally, amongst the four models, DeepDeath is performing the best due to its ability in learning sequential data. A noteworthy point is that, as opposed to the baselines, we do not need to hand-craft features using ad-hoc rules proposed by an expert. Instead, without having any intuition about the nature of the data, the LSTM network can learn the rules of its own. In fact, this observation has been confirmed in numerous other applications and is a key factor leading to the versatility of deep models, and LSTMs in particular.

One important caveat of deep pipelines is the lack of interpretability. As opposed to many classical models such decision trees, deep models do not generate understandable rules that human can utilize to generalize the concept. In light of this shortcoming, we were interested to see how interpretable are the intermediate features that DeapDeath generate. To this end, upon training the model, we removed the SoftMax and the drop-out layers and applied the model to the test set and stored the intermediate features that the fully connected layer generates per each sample. Then, we marked each death either as a death caused by an infectious or parasitic disease (if the ICD-10 code for the underlying cause starts with "A" or "B") or a non-infectious cause (i.e. all others). Next, we used t-SNE [21] a popular visualization technique that maintains the locality of the samples, to map the features into a 2D plane. Our understanding is that the pathology and disease progression patterns of infectious and parasitic diseases are often different from other diseases. Figure 4 depicts the resulting visualization. Interestingly, we observe that there is a meaningful pattern observed among the samples in each group suggesting that intermediate features that are generated from the raw features (i.e. the multiple cause trajectories) can be visualized and interpreted by human experts.

TABLE 1: PERFORMANCE COMPARISION BETWEEN PREDICTIONS MADE BY THE BASELINES AND THE DEEPDEATH ON 67 LABELS

| Random Forest on Hadoop | | | DeepDeath |
|---|---|---|---|
| *UniGram (5000)* | *BiGram (5000)* | *Uni+BiGram (10000)* | |

| ACC (67 classes) | 31.92% | 22.11% | 34.48% | **36.98%** |

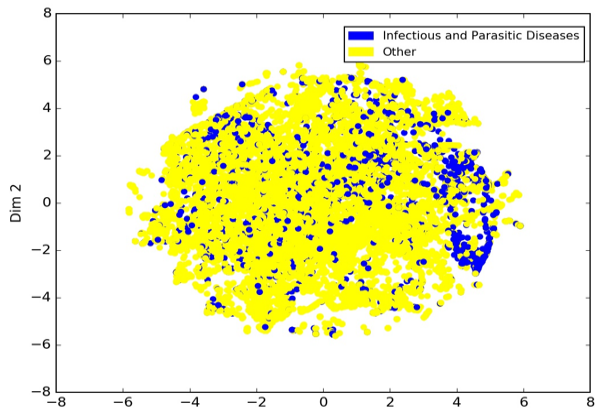

Figure 4. Visualization of intermediate features generated by DeepDeath for deaths with infectious and parasitic causes vs. the others.

## IV. CONCLUSION AND DISCUSSION

In this study, we proposed two classes of models for analyzing large scale mortality data. We showed that both classes significantly perform better than the random classifier. Moreover, through addition of bi-gram features to uni-gram features, we showed that if temporal aspects of the input data captured properly, an improvement on the classification task can be achieved, a fact that motivated us to design a model based on the deep long short-term memory networks. One of the active research directions in the field of deep learning is finding effective ways to interpret what deep models learn. In this study we used visualization of the intermediate features as a first step to solve this problem and we showed that meaningful clusters of intermediate features may help understanding what salient features the deep models have learned.

This study sets the stage for a comprehensive decision support system that can assist physicians and practitioners in filling out the death certificate correctly. As a future work, we are interested to develop generative (as opposed to the current discriminative) models that can suggest the most probably correct way to fill out the death certificate forms given the available data that physician/practitioner has in hand.

ACKNOWLEDGMENT

The authors would like to thank Paula A. Braun for her contribution to this research.